\documentclass{article}
\usepackage{spconf,amsmath,graphicx}

\usepackage{epsfig}
\usepackage{amssymb}

\usepackage[utf8]{inputenc} % allow utf-8 input
\usepackage[T1]{fontenc}    % use 8-bit T1 fonts
\usepackage{url}            % simple URL typesetting
\usepackage{booktabs}       % professional-quality tables
\usepackage{amsfonts}       % blackboard math symbols
\usepackage{nicefrac}       % compact symbols for 1/2, etc.
\usepackage{microtype}      % microtypography
\usepackage{tabularx, multirow, graphics}
\usepackage{times,helvet,courier,graphicx,booktabs,amsmath,caption,subcaption,mathtools,listings,xcolor}
\usepackage[shortlabels]{enumitem}

\usepackage{algpseudocode, comment}
\usepackage{bbold}

\usepackage[ruled]{algorithm2e} % For algorithms
\newenvironment{heuristic}[1][htb]
  {% Update algorithm name
   \begin{algorithm}[#1]%
  }{\end{algorithm}}

\def\bx{{\boldsymbol x}}
\def\by{{\boldsymbol y}}

\def\bp{{\boldsymbol p}}

\def\be{{\boldsymbol e}}
\def\bq{{\boldsymbol q}}

\def\bzero{{\boldsymbol 0}}
\def\btheta{\boldsymbol \theta}

\def\EB0{{\tt EfficientNet-B0}}
\def\SNNEB0{{\tt SNN-EB0}}
\def\RESNET50{{\tt ResNet-50}}

\newcommand\T{\rule{0pt}{2ex}}     % Top strut
\newcommand\B{\rule[-1ex]{0pt}{0pt}} % Bottom strut

\title{KD-FixMatch: Knowledge Distillation Siamese Neural Networks}
\name{
    Chien-Chih Wang, Shaoyuan Xu, Jinmiao Fu, Yang Liu, Bryan Wang
}
\address{
    \{ccwang, shaoyux, jinmiaof, yliuu, brywan\}@amazon.com \\
    Amazon Inc.
}

\begin{document}
%\ninept
%
\maketitle

\begin{abstract}
	%Learning from large-scale high-quality labeled data is a crucial factor in the success of deep neural networks.
	%However, data labeling is a time-consuming and unscalable task, which often results in the shortage of labeled data.
	%To address this challenge, semi-supervised learning (SSL) is employed to leverage additional unlabeled data.
	%One of the popular SSL algorithms, {\sl FixMatch} \cite{KS20a}, is based on a Siamese Neural Network (SNN) that trains identical weight-sharing teacher and student networks simultaneously.
	%However, it is prone to performance degradation when the pseudo labels are heavily noisy in the early training stage.
	%To overcome this limitation, we propose {\sl KD-FixMatch}, an algorithm that combines sequential and simultaneous teacher-student training.
	%We first train an outer SNN using labeled and unlabeled data.
	%The well-trained outer SNN then distills knowledge by generating pseudo labels for the unlabeled data, from which a subset of trusted pseudo labels is carefully selected using deep embedding clustering and high-confidence sampling.
	%Finally, an inner SNN is trained with not only labeled and unlabeled data but also the selected unlabeled data with trusted pseudo labels.
	Semi-supervised learning (SSL) has become a crucial approach in deep learning as a way to address the challenge of limited labeled data.
    The success of deep neural networks heavily relies on the availability of large-scale high-quality labeled data.
    However, the process of data labeling is time-consuming and unscalable, leading to shortages in labeled data.
    SSL aims to tackle this problem by leveraging additional unlabeled data in the training process.
    One of the popular SSL algorithms, {\sl FixMatch} \cite{KS20a}, trains identical weight-sharing teacher and student networks simultaneously using a siamese neural network (SNN).
    However, it is prone to performance degradation when the pseudo labels are heavily noisy in the early training stage.
    We present {\sl KD-FixMatch}, a novel SSL algorithm that addresses the limitations of {\sl FixMatch} by incorporating knowledge distillation.
    The algorithm utilizes a combination of sequential and simultaneous training of SNNs to enhance performance and reduce performance degradation.
    Firstly, an outer SNN is trained using labeled and unlabeled data. After that, the network of the well-trained outer SNN generates pseudo labels for the unlabeled data, from which
    a subset of unlabeled data with trusted pseudo labels is then carefully created through high-confidence sampling and deep embedding clustering.
    %The unlabeled data with pseudo labels is then carefully selected through high-confidence sampling and deep embedding clustering, creating a subset of unlabeled data with trusted pseudo labels.
    Finally, an inner SNN is trained with the labeled data, the unlabeled data, and the subset of unlabeled data with trusted pseudo labels.
	Experiments on four public data sets demonstrate that {\sl KD-FixMatch} outperforms {\sl FixMatch} in all cases.
	Our results indicate that {\sl KD-FixMatch} has a better training starting point that leads to improved model performance compared to {\sl FixMatch}.
\end{abstract}
\begin{keywords}
semi-supervised learning, knowledge distillation, siamese neural networks, high-confidence sampling, deep embedding clustering
\end{keywords}
\section{Introduction}
\label{sec:intro}

%Large-scale high-quality labeled data is the key to tremendous success of supervised deep learning models in many fields \cite{RG15a, SR15a, JC17a, JD18a}.
Large-scale high-quality labeled data is the key to the tremendous success of supervised deep learning models in many fields.
However, collecting labeled data remains a significant challenge, as the process of data labeling is time-consuming and resource-intensive.
This is particularly evident in tasks such as detecting product image defects in e-commerce, where high-quality labeled data is essential for accurate predictions and customer satisfaction.
The numbers of defective and non-defective images are sometimes extremely unbalanced and thus
it is necessary to label a large number of randomly selected images in order to collect sufficient high-quality labeled images for supervised learning.

\begin{comment}
However, limited labeled data is still a fundamental challenge to train deep neural networks (DNNs) because data labeling is exceedingly resource consuming.
For example, building a supervised deep learning model for detecting product image defects on e-commerce websites,
which has significant impacts on customers' decisions, requires a lot of high-quality labeled data.
Unfortunately, the data for defective and non-defective images is extremely unbalanced and
we need to label a large number of randomly selected images in order to collect enough defective labeled images for supervised learning.
\end{comment}

%Comparing to labeled data, it is more convenient to collect large-scale unlabeled data.
%For example, the product images uploaded by the sellers can be directly used as unlabeled data without labeling.
%Fortunately,
%SSL utilizes the information from both labeled and unlabeled data in the learning process, which can either improve the model performance with same amount of labeled data,
%or reach on-par results with much less labeled data.

Recently, \cite{KS20a} came up with a widely used SSL algorithm called {\sl FixMatch}, which trains an SNN with limited labeled and extra unlabeled data,
and achieved significant improvement comparing to traditional supervised learning methods.
However, despite the superiority and success of {\sl FixMatch}, it has a noticeable disadvantage.
It trains the teacher and the student in an SNN simultaneously and since the pseudo labels generated by the teacher in the early stage may not be correct, using them directly may introduce large amount of label noise.
Therefore, we propose a modified {\sl FixMatch} called {\sl KD-FixMatch}, which trains an outer and an inner SNN sequentially.
Our proposed method improves upon {\sl FixMatch} because an outer SNN will be first trained with labeled and unlabeled data and thus
the percentage of label noise in the pseudo labels generated by the network of the well-trained outer SNN would be lower than that of the teacher in {\sl FixMatch}, especially in the early stage.
Our main contributions are summarized as follows:
\vspace{-2.0mm}
\begin{enumerate}[1.]
    %\vspace{3.0mm}
    \item
        In {\sl KD-FixMatch}, the outer SNN is first well-trained and its network serves as a teacher for the inner SNN.
        Thus, we have a better training starting point than direcly applying {\sl FixMatch} algorithm to the inner SNN.
    \vspace{-2.0mm}
    \item
        Unlike self-training \cite{DY95a,DM06a} or Noisy Student \cite{QX20a}, where its neural network is re-trained multiple times when its pseudo labels are required to be updated,
        we only need to train each of the outer and inner SNN once.
    \vspace{-2.0mm}
    \item
        {\sl KD-FixMatch} outperforms {\sl FixMatch} in our experiments
        although the time complexity of {\sl KD-FixMatch} is more than two times that of {\sl FixMatch}.
        %\footnote{
        %    For {\sl KD-FixMatch}, the three main procedures are to firstly train an outer SNN with labeled and unlabeled data,
        %    secondly generate pseudo labels for the unlabeled data using the well-trained outer SNN, and lastly train an inner SNN using
        %    not only the labeled and unlabeled data but also the unlabeled data with trusted pseudo labels.
        %}
    \vspace{-2.0mm}
\end{enumerate}

\section{RELATION TO PRIOR WORK}
\label{sec:related-works}

%SSL has recently become a popular research topic \cite{KS20a,QX20a,AO18a,DB19a,HP21a,BZ21a,KN21a,DL22a,WC22a}
\begin{comment}
SSL has recently become a popular research topic \cite{KS20a,QX20a,AO18a,DB19a,BZ21a,WC22a}
due to its capability to significantly improve the model performance by leveraging a large-scale unlabeled data when the labeled data is limited.
\end{comment}
%However, SSL can only achieve satisfactory results when the distributions of the labeled and unlabeled data are similar \cite{AO18a,XY21a}.
%which often observed in the well-organized and well-annotated public data sets.
%For real-world applications, SSL model may not be better, even worse, than the model trained only with the labeled data by supervised learning methods
%if the distributions of the labeled and unlabeled data are very different.
SSL has gained significant attention in recent years as a method to overcome the challenge of limited labeled data.
%By incorporating large-scale unlabeled data, SSL has shown to enhance the performance of deep neural networks, as demonstrated in various studies \cite{KS20a,QX20a,AO18a,DB19a,BZ21a,WC22a}.
By incorporating large-scale unlabeled data, SSL has shown to enhance the performance of deep neural networks, as demonstrated in various studies \cite{KS20a,QX20a,DB19a,BZ21a,WC22a}.

%\subsection{Self-training}
%\label{subsec:self-training}

%Self-training, which is also called self-labeling or self-teaching, is a straightforward SSL algorithm.
%Firstly, we train a model based on the initial labeled data.
%Secondly, the well-trained model generates pseudo labels for the unlabeled data. After that, we only pick the unlabeled data with highly confident pseudo labels and add them to the initial labeled data to form a new labeled data.
%Lastly, we use the new labeled data to train a new model and perform this procedure iteratively.
%However, one of the drawbacks of self-training is time-consuming.

\subsection{Teacher-Student Sequential Training}
\label{subsec:KD}

Knowledge distillation \cite{GH15a} is a well-known model compression technique that transfers knowledge from a pre-trained, larger teacher model to a smaller student model.
By using the teacher to generate pseudo labels for the unlabeled data, the student trained on both labeled data and pseudo-labeled data performs better than when trained solely on the limited labeled data \cite{GH15a}.

Noisy Student \cite{QX20a} takes inspiration from knowledge distillation and uses a student model that is equal to or larger than the teacher model.
This allows for the inclusion of noise-inducing techniques such as dropout, stochastic depth, and data augmentation in the student's training, resulting in better generalization compared to the teacher \cite{QX20a}.
In Noisy Student, once the student is found to be better than the teacher, the pseudo labels are updated by the current student,
and a new student is re-initialized. This procedure is then repeated several times.
However, this approach requires waiting for the teacher to be well-trained before generating highly confident pseudo labels for the student's training.

\subsection{Teacher-Student Simultaneous Training}
\label{subsec:TS-simultaneous}

\begin{figure}[t]
    \begin{center}
    \includegraphics[scale=0.21]{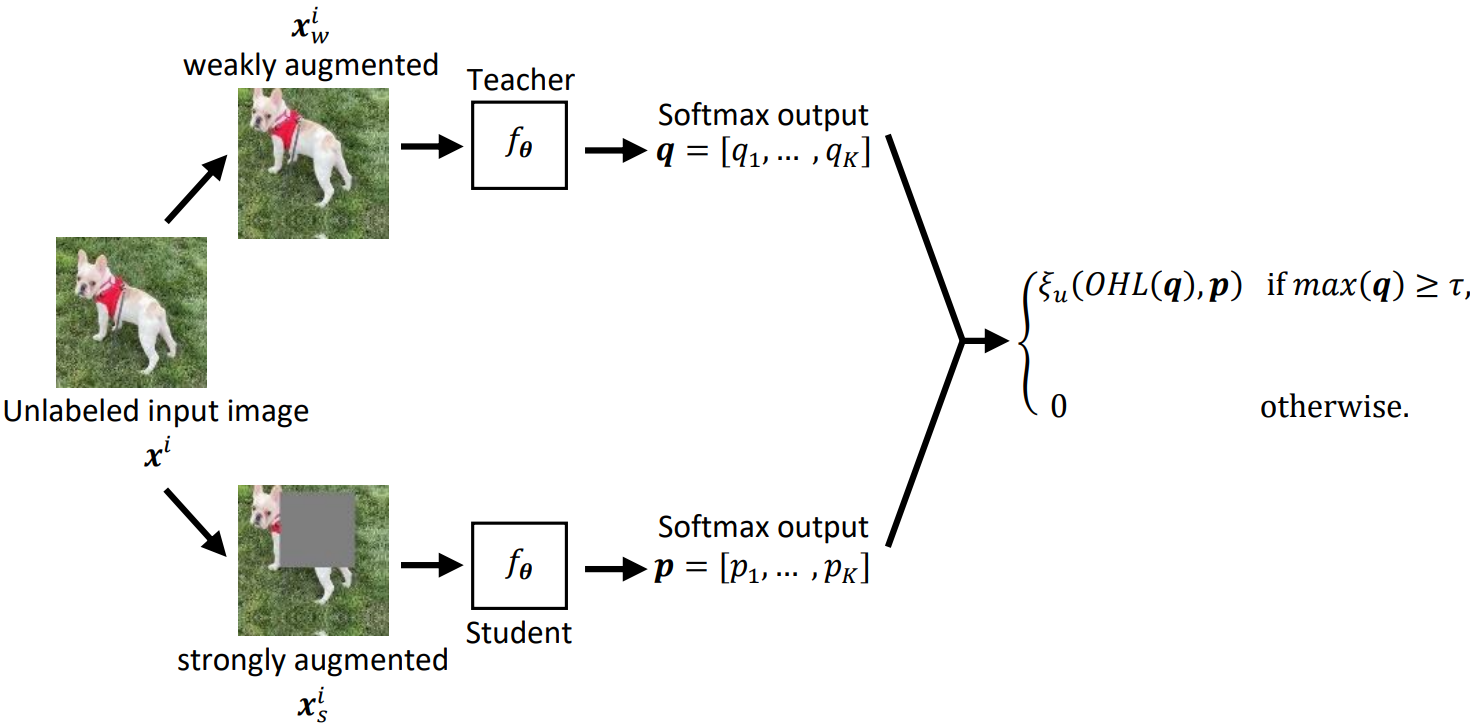}
    \end{center}
    \vspace{-5mm}
    \caption{An example of SNN for {\sl FixMatch}.
        %A standard neural network, $f_{\btheta}$, with the softmax output layer is provided.
        %For an input image, we first applied weakly and strongly augmentations on the input image, respectively.
        %Then, the weakly and strongly augmented images are feed into $f_{\btheta}$, respectively.
        %When the input of $f_{\btheta}$ is the weakly augmented image, $f_{\btheta}$ is considered as a teacher
        %and its softmax output is the pseudo label of the input image.
        %Conversely, $f_{\btheta}$ is considered as a student when the input of $f_{\btheta}$ is the strongly augmented image.
        %In order to reduce noise label problem, {\sl FixMatch} only trust the maximum value of the pseudo label that is greater than or equal to a pre-defined threshold, $\tau$.
        %Lastly, when the trusted pseudo label is selected, the loss function, $\xi_{\text{in}}$, calculates the distance between the softmax output of the strongly augmented image and the trusted pseudo label and
        %back-propagation is applied to update the weights in $f_{\btheta}$.
    }
\label{fig:SNN-explain}
\end{figure}

Besides the aforementioned sequential methods, the simultaneous training method is also an alternative option \cite{KS20a,HP21a,AT17a}.
We take {\sl FixMatch} \cite{KS20a} as an example.
The core neural network architecture for {\sl FixMatch} is SNN which consists of two identical weight-sharing classifier networks, a teacher and a student.
See Figure \ref{fig:SNN-explain}.

\par Assume that $f_{\btheta}$ is a standard neural network with softmax output layer,
$\btheta \in \Re^n$ is a long vector, $\ell$ is the number of training examples, $K$ is the number of the classes, and $n$ is the total number of the parameters of $f_{\btheta}$.
We define the optimization problem for {\sl FixMatch} to be
\begin{equation}
\label{fixmatch-obj}
    \min_{\btheta}\ f(\btheta), \text{where}\ f(\btheta) = 
    \frac{1}{2C} \btheta^T \btheta + 
    \frac{1}{\ell} \sum_{i=1}^\ell \xi(\btheta;\bx^i,\by^i)
\end{equation}
$C > 0$ is the parameter to avoid overfitting by regularization, $\bx^i$ is the $i$th input image, and $\by^i$ is the label vector of $\bx^i$.
In addition, if $\bx^i$ belongs to the $s$th class, the label vector, $\by^i$, is $[\underbrace{0, \ldots, 0}_{s-1}, 1, 0, \ldots, 0]^T \in \Re^K$.
%\begin{equation*}
%    \by^i = [\underbrace{0, \ldots, 0}_{s-1}, 1, 0, \ldots, 0]^T \in \Re^K.
%\end{equation*}
However, training examples for SSL consists of both a set of the labeled data, $S_l$, and a set of the unlabeled data, $S_u$.
Label vectors for the unlabeled data are not available.
Therefore, {\sl FixMatch} uses teacher inside SNN to generate pseudo labels as label vectors for $\bx^i, i \in S_u$  
and then the loss function, $\xi$, in \eqref{fixmatch-obj} can be defined as follows:
\begin{equation}
\label{fixmatch-loss}
    \xi(\btheta;\bx^i,\by^i) = 
    \begin{cases}
        \xi_l\left(\by^i, f_{\btheta}(\bx^i_{w})\right) & \text{if}\ \ i \in S_{l},\\
        \lambda_u \xi_u(\hat{\by}^i, f_{\btheta}(\bx^i_{s})) & \text{if}\ \ i \in S_{u},
    \end{cases}
    %\left(
    %\underbrace{\sum_{i \in B_{l}} \xi_l\left(\by^i, f_{\btheta}(\bx^i_{w})\right)}_{\text{labeled loss}} 
    %+ \underbrace{\sum_{j \in B_{u}} \xi_u\left(\overbrace{f_{\btheta}(\bx^j_{w})}^{\text{teacher}}, \overbrace{f_{\btheta}(\bx^j_{s})}^{\text{student}}\right)}_{\text{unlabeled loss}}
    %\right),
\end{equation}
where $\hat{\by}^i$ is the pseudo label vector of $\bx^i, i \in S_u$, $\lambda_u$ is a pre-defined parameter, $\bx^i_w$ is the weakly augmented image of $\bx^i$, $\bx^i_s$ is the strongly augmented image of $\bx^i$,
$\xi_l$ is a loss function for the labeled data, and $\xi_u$ is a loss function for the unlabeled data.

\par In \cite{KS20a}, $\xi_l$ in \eqref{fixmatch-loss} is the standard cross-entropy (CE) function to measure the difference between $f_{\btheta}(\bx^i_{w})$ and its label vector $\by^i$.
For $\xi_u$ in \eqref{fixmatch-loss},
when the input of $f_{\btheta}$ is $\bx^i_w$, $f_{\btheta}$ is considered as the teacher and
$f_{\btheta}(\bx^i_{w})$ is $\hat{\by}^i$ of $\bx^i$.
Conversely, $f_{\btheta}$ is considered as the student when the input of $f_{\btheta}$ is $\bx^i_s$.
However, in order to reduce label noise, we only choose the highly confident pseudo labels whose maximum value are greater than or equal to a pre-defined threshold $\tau > 0$
and thus $\xi_u$ in \eqref{fixmatch-loss} can be defined as follows:
\begin{equation}
\label{fixmatch-filter}
    %\xi_u\left(f_{\btheta}(\bx^j_{w}), f_{\btheta}(\bx^j_{s})\right) =
    \mathbb{1}(\max(\hat{\by}^i) \geq \tau )
      \xi_{u}(\text{OHL}(\hat{\by}^i), f_{\btheta}(\bx^i_{s})),
\end{equation}
where $\mathbb{1}$ is an indicator function and OHL is a function mapping a softmax pseudo label vector to its one-hot label vector.
After the loss function, $\xi$, in \eqref{fixmatch-obj} is prepared, back-propagation is applied to update the weights in $f_{\btheta}$ and thus
both teacher and student are updated simultaneously by minimizing \eqref{fixmatch-obj}.

For the method where teacher and student are trained simultaneously, there is no need to wait for the teacher's training procedure to complete before the student's can begin.
However, in the early stage, heavy label noise or insufficient correct pseudo labels
generated by the immature teacher may affect the generalization ability of the student model.

\section{Proposed Algorithm}
\label{sec:proposed-alg}

We propose an algorithm called {\sl KD-FixMatch}, which contains an outer SNN, $f^{\text{outer}}$, and an inner SNN, $f^{\text{inner}}$.
The goal is not only to use teacher-student sequential training to fix label noise problem in the early stage,
but also to incorporate teacher-student simultaneous training such as {\sl FixMatch}.
%We first initialize an outer SNN and train it with the labeled and unlabeled data by using back-propagation to solve \eqref{fixmatch-obj}.
%Then, the well-trained outer SNN generates the pseudo labels for the unlabeled data before inner SNN training.
%However, the well-trained outer SNN is not perfect. Therefore, we provide a strategy
%to choose a trusted subset from the pseudo labels in order to reduce label noise.

%Furthermore, during the inner SNN's training, the conflict may arise when both the well-trained outer SNN and inner SNN provide their own trusted pseudo labels for the same unlabeled image.
%Intuitively, our inner SNN has better generalization ability than the well-trained outer SNN
%because it has the additional unlabeled data with trusted pseudo labels in the early stage.
%Therefore, in Section \ref{subsec:heuristic-teacher-student}, we propose heuristics to resolve the conflict.
%In addition, some past SSL works \cite{WC22a} mentioned that robust loss functions may be helpful in case of noisy pseudo labels and we introduce them in Section \ref{subsec:robust-loss}.
%Finally, we summarize our proposed algorithm in Section \ref{subsec:dual-TS}.

\subsection{Trusted Pseudo Label Selection}
\label{subsec:outer-trust-selection}

After the outer SNN is well-trained, it generates pseudo labels for the unlabeled data.
Due to the imperfection of the outer SNN, we need to choose a subset of trusted pseudo labels to reduce label noise.
Otherwise, it may affect inner SNN's model performance.

Similar to \eqref{fixmatch-filter}, we first choose a pre-defined threshold, $\tau^{\text{select}}$, and choose the pseudo labels whose maximum value is greater than or equal to $\tau^{\text{select}}$.
Then, we extract the latent representations\footnote{
The representations are derived from the layer before the last layer.
}
of the unlabeled data which has passed the $\tau^{\text{select}}$ selection.
We do a deep embedding clustering \cite{JX16a, XY21a} on the representations and
choose the unlabeled data whose clustering results are consistent with their predicted class.\footnote{
    The predicted class is the index of the maximum value of $\hat{\by}^i$.    
}
Finally, the indices of the chosen unlabeled data form $T_u$.

\subsection{Merging Conflict Pseudo Labels}
\label{subsec:heuristic-teacher-student}

From Section \ref{subsec:outer-trust-selection}, $T_u$ is determined.
However, our inner SNN also generates pseudo labels for the unlabeled data during its training process.
Following \eqref{fixmatch-filter}, we choose the pseudo labels whose maximum value is greater than or equal to
a pre-defined threshold, $\tau^{\text{inner}} > 0$, to be inner SNN's trusted pseudo labels.
Therefore, conflicts may arise when both the outer and inner SNN provide their own trusted pseudo labels for the same image.
%In some past teacher-student works \cite{DM06a,QX20a}, pseudo labels are re-generated when their student model is better than their teacher model.
%After that, a new student model will be re-initialized and then re-trained with the newly generated pseudo labels.
%This is deemed reasonable, as it is expected that we can obtain a better resulting model if we have better pseudo labels.
%Unfortunately, it is time-consuming and non-practical.
Heuristic \ref{lst:heuristic} is proposed to determine the pseudo label vector, $\hat{\by}^i$, in \eqref{fixmatch-filter} for the unlabeled data which has conflict pusedo labels.
The main idea is that
the inner SNN intuitively has better generalization ability than the outer SNN since it has more reliable pseudo labels in the early stage.
%Comparing to the past works, by using Heuristic \ref{lst:heuristic}, our proposed algorithm only needs to run inner SNN training once.
\begin{heuristic}[t]
    \uIf{$\max\left(f^{\text{inner}}_{\btheta}(\bx^i_{w})\right) \geq \tau^{\text{inner}}$}{
        $\hat{\by}^i \leftarrow f^{\text{inner}}_{\btheta}(\bx^i_{w})$  
  }
    \uElseIf{$i \in T_u$}{
        $\hat{\by}^i \leftarrow \by^{i,\text{outer}}$
  }
  \Else{
        $\hat{\by}^i \leftarrow \bzero$
  }
\caption{Merging Conflict Pseudo Labels}
\label{lst:heuristic}
\end{heuristic}

\subsection{Robust Loss Functions}
\label{subsec:robust-loss}

Label noise is one of the major problems for SSL since teacher model is not perfect.
In other words, there exist noisy labels among the pseudo labels generated by the teacher.
\cite{WC22a} mentioned that robust loss functions may be helpful in case of noisy pseudo labels.
The widely used loss functions in robust learning is symmetric loss functions \cite{NM13a,AG17a} which has been proven to be robust to noisy labels.
However, there are some constraints for symmetric loss functions.

Assume that $f$ is a standard neural network with softmax output layer.
From \cite{AG17a}, a loss function $\xi$ is called symmetric if it satisfies
%\begin{equation*}
%\label{robust-loss-def}
$\sum_{k=1}^K \xi(f(\bx), \be_k) = M,\ \forall \bx \in X,\ \forall f,$
%\end{equation*}
where $K$ is the number of classes, $X$ is the feature space, 
$f$ is a function mapping an input $\bx$ to a softmax output $\by$, $M$ is a constant value, and
%\begin{equation*}
$\be_k = [\underbrace{0, \ldots, 0}_{k-1}, 1, 0, \ldots, 0]^T \in \Re^K$.
%\end{equation*}

Mean absolute error (MAE) \cite{CJW85a}, symmetric cross-entropy (SCE) \cite{YW19a}, and normalized cross-entropy (NCE) \cite{XM20a} are examples of robust loss functions.

\subsection{Knowledge Distillation Siamese Neural Networks}
\label{subsec:dual-TS}

%After $f^{\text{inner}}$, $T_u$, $\xi^{\text{inner}}_l$, $\xi^{\text{inner}}_u$, $\tau^{\text{inner}}$ are determined, 
We initialize an inner SNN and then train it with not only the labeled and unlabeled data but also the unlabeled data with trusted pseudo labels, $(\bx^i, \hat{\by}^i), i \in T_u$.
For the inner SNN, the optimization problem and loss functions are the same as \eqref{fixmatch-obj}, \eqref{fixmatch-loss}, and \eqref{fixmatch-filter} except that
we substitute $f$, $C$, $\xi_l$, $\lambda_u$, $\xi_u$, and $\tau$  with $f^{\text{inner}}$, $C^{\text{inner}}$, $\xi^{\text{inner}}_l$, $\lambda^{\text{inner}}_u$, $\xi^{\text{inner}}_u$, and $\tau^{\text{inner}}$, respectively.
%In addition, we use Heuristic \ref{lst:heuristic} to determine $\hat{\by}^i, i \in T_u$ for \eqref{fixmatch-loss} and \eqref{fixmatch-filter}.
%Furthermore, we can not only use CE function but also robust loss functions mentioned in Section \ref{subsec:robust-loss} \cite{WC22a} for $\xi^{\text{inner}}_u$.
A summary of our proposed algorithm is in Algorithm \ref{alg:os-fixmatch}.

\setcounter{algocf}{0}
\begin{algorithm}[t]
%\SetAlgoNoLine
    Given a labeled data, $S_l$, and an unlabeled data, $S_u$\;
    1: Initialize an outer SNN, $f^{\text{outer}}$, and train it with $S_l$ and $S_u$ by using back-propagation to
    solve \eqref{fixmatch-obj}\;
    2: Generate pseudo labels for $S_u$ using the well-trained $f^{\text{outer}}$\;
    3: Select a trusted index subset $T_u$ from $S_u$\; 
    4: Choose CE or a robust loss function to be $\xi^{\text{inner}}_u$\;
    5: Initialize an inner SNN, $f^{\text{inner}}$, and train it with not only the labeled and unlabeled data but also the unlabeled data with trusted pseudo labels, $(\bx^i, \hat{\by}^i), i \in T_u$\;
    6: Return the well-trained $f^{\text{inner}}$\;
    \caption{KD-FixMatch}
\label{alg:os-fixmatch}
\end{algorithm}

\section{Experiments}
\label{sec:exps}

The main objective in this section is to compare our proposed algorithm with other methods.
For a fair comparison, we implement all of the methods using {\sl Tensorflow 2.3.0}.
%In order to check the correctness of our implementation,
%we run the experiments in \cite{KS20a} using our implementation and derive the similar or competitive results. See Appendix A.
\EB0 \cite{MT19a} pre-trained on {\sf ImageNet}\footnote{
Pre-trained models are derived from \url{https://www.tensorflow.org/api_docs/python/tf/keras/applications}.
}
is chosen for all of the experiments and
\SNNEB0 is an SNN which consists of two identical weight-sharing \EB0.
We compare the following five methods:
\footnote{
    Due to the space limitation, for robust loss functions, we only use SCE for our experiments.
    The formula of SCE is $\alpha \times \text{CE} + \beta \times \text{RCE}$ and
    RCE stands for reverse cross-entropy.
    Assume that $H(\bq, \bp)$ is a standard CE function given two distributions, $\bq$ (ground truth distribution) and $\bp$ (distribution from the softmax neural network output),
    RCE can be defined as: $H(\bp, \bq)$.
}
%and the details of the used parameters are shown in Appendix F.
(a) {\sl Baseline}: \EB0 is trained with only the labeled data by solving \eqref{fixmatch-obj} with CE loss function;
(b) {\sl FixMatch}: \SNNEB0 is trained with the labeled and unlabeled data by solving \eqref{fixmatch-obj} with \eqref{fixmatch-loss} and \eqref{fixmatch-filter} and both $\xi_l$ and $\xi_u$ are CE functions;
(c) {\sl KD-FixMatch-CE}: an outer and an inner \SNNEB0 are trained using Algorithm \ref{alg:os-fixmatch} with the labeled and unlabeled data 
        and both $\xi_l$ and $\xi_u$ for outer and inner \SNNEB0 are CE functions;
(d) {\sl KD-FixMatch-SCE-1.0-0.01}: the same as {\sl KD-FixMatch-CE} except that
        $\xi_u$ for inner \SNNEB0 is an SCE function with $\alpha = 1.0$ and $\beta = 0.01$; and
(e) {\sl KD-FixMatch-SCE-1.0-0.1}: the same as {\sl KD-FixMatch-CE} except that
        $\xi_u$ for inner \SNNEB0 is an SCE function with $\alpha = 1.0$ and $\beta = 0.1$.
For the labeled training data in all of our experiments, we follow the sampling strategy mentioned in \cite{KS20a} to choose an equal number of images from each class in order to avoid model bias.
%An example of such bias can be that our resulting model is prone to predict the negative class if the number of the negative training data is much larger than the positive training data.

For all the experiments, Adam \cite{DK14a} optimizer is used and the exponential decay rates for $1$st and $2$nd moment estimates are $0.9$ and $0.999$, respectively.
Batch size is set to $128$.\footnote{
Except for {\sl Baseline}, $64$ labeled and $64$ unlabeled images are in a batch.
}
The initial learning rate is $7\mathrm{e}{-5}$ and the learning rate scheduler mentioned in Section $5.2$ of \cite{MT19a} is applied.
We set $1/C$ in \eqref{fixmatch-obj} to be $5\mathrm{e}{-4}$ except that $1\mathrm{e}{-3}$ for {\sf CIFAR100}.
For weak augmentation, we follow the method in Section 2.3 of \cite{KS20a}.
For strong augmentation, we follow RandAugment implementation in \cite{EC20a}.
%\footnote{
%    See \url{https://github.com/kekmodel/FixMatch-pytorch/blob/98cedd6ffca4813fe6d744f695bee52beaf0faf7/dataset/cifar.py} and \url{https://github.com/kekmodel/FixMatch-pytorch/blob/master/dataset/randaugment.py}.
%}
Following \cite{KS20a}, we set $\tau$ in \eqref{fixmatch-filter} to be $0.95$ and $\lambda_u$ be $1$.
Also, we set $\tau^{\text{select}}$ and $\tau^{\text{inner}}$ to be $0.80$ and $0.95$, respectively.
For inference, we follow \cite{KS20a} to use the maintained exponential moving average of the trained parameters.
The decay is set to be $0.9999$ and num\_updates be $10$.
\footnote{
    See \url{https://github.com/petewarden/tensorflow_makefile/blob/master/tensorflow/python/training/moving_averages.py}.
}

\subsection{Experiments on Four Public Data Sets}
\label{subsec:exp-public}

In this subsection, we conduct our experiments on four public data sets: {\sf SVHN} \cite{YN11a}, {\sf CIFAR10/100} \cite{AK09a}, and {\sf FOOD101} \cite{LB14a}.
In order to mimic the practical use, we randomly split the training data from their official websites into original training data ($80\%$) and original validation data ($20\%$) for our experiments.
The unlabeled data from the official websites is not used. Instead, we follow \cite{KS20a} to ignore the labels in the original training data and deem it as unlabeled data.
All of these four data sets are publicly available
%\footnote{
%{\sf SVHN}: \url{http://ufldl.stanford.edu/housenumbers/},
%{\sf CIFAR10/100}: \url{https://www.cs.toronto.edu/~kriz/cifar.html}, and
%{\sf FOOD101}: \url{https://data.vision.ee.ethz.ch/cvl/datasets_extra/food-101/}.
%}
and the summary is in Table \ref{table:data-statistic}.
\begin{table}[t]
    \centering
    \resizebox{0.85\columnwidth}{!}{%
        \begin{tabular}{l c c c c}
        \hline
            & \#class & \#train & \#validation & \#test \T\B \\
        \hline
            {\sf CIFAR10} & 10 & 40,000 & 10,000 & 10,000 \T\B\\
        \hline
            {\sf SVHN} & 10 & 58,606 & 14,651 & 26,032 \T\B\\
        \hline
            {\sf CIFAR100} & 100 & 40,000 & 10,000 & 10,000 \T\B\\
        \hline
            {\sf FOOD101} & 101 & 60,600 & 15,150 & 25,250 \T\B\\
        \hline
        \end{tabular}
    }
    \caption{Summary of four public data sets.
        \#class is the number of classes, \#train
        is the number of the train data, \#validation is the number of the validation data,
        \#test is the number of the test data.
    }
\label{table:data-statistic}
\end{table}

%Assume that we need to sample a total of $\ell$ training images from the original training data.
%Following \cite{KS20a}, from the original training data, we randomly sample $\floor*{\frac{\ell}{K}}$ images from each of the $K$ classes.\footnote{
%    When $\ell$ is not divisible by $K$, if the remainder is $r$, we randomly sample $r$ images from the non-selected original training data.
%}
%We take {\sf CIFAR10} using $400$ labeled images as an example. The number of total labeled images is $400$ and we randomly sample $40$ images from each of the $10$ classes.
From Table \ref{table:exp0}, we observe that:
(a) {\sl FixMatch} is better than {\sl Baseline} for all the cases.
This is as expected because, comparing to {\sl Baseline}, {\sl FixMatch} leverages an additional unlabeled data;
(b) {\sl KD-FixMatch} is better than {\sl FixMatch} for all the cases.
The reason is that {\sl KD-FixMatch} has a better starting point than {\sl FixMatch};
(c) Comparing to {\sl FixMatch}, the more labeled data we use, the less improvements we have for {\sl KD-FixMatch}.
The possible reason is that when {\sl FixMatch} has a large enough initial labeled data it has a good starting point to reach the result competitive to {\sl KD-FixMatch}; and
(d) {\sl KD-FixMatch-SCE} has slightly better model performance than {\sl KD-FixMatch-CE} in most of the cases even without grid search for the two parameters, $(\alpha, \beta)$.
The possible reason is that the robust loss function, SCE, is applied.
\begin{table}[t]
% CIFAR10
    \begin{center}
    \resizebox{1.0\columnwidth}{!}{%
    \begin{tabular}{l | c c c c}
    \hline
      & \multicolumn{1}{c}{$400$ labels}
      & \multicolumn{1}{c}{$1,000$ labels}
      & \multicolumn{1}{c}{$2,000$ labels}
      & \multicolumn{1}{c}{$4,000$ labels}
        \T\B\\
    \hline
        \multicolumn{1}{l|}{{\sl Baseline}}
        %& \% & \% & \% & \%
        & $33.47 \pm 4.62\%$
        & $44.18 \pm 4.40\%$
        & $82.62 \pm 1.90\%$
        & $93.37 \pm 0.24\%$
        \T\B\\
    \hline
        \multicolumn{1}{l|}{{\sl FixMatch}}
        %& \% & \% & \% & \%
        & $84.96 \pm 0.75\%$
        & $91.85 \pm 0.58\%$
        & $94.25 \pm 0.47\%$
        & $95.69 \pm 0.15\%$
        \T\B\\
    \hline
        \multicolumn{1}{l|}{{\sl KD-FixMatch-CE}}
        %& \% & \% & \% & \%
        & $87.40 \pm 0.53\%$
        & $92.92 \pm 0.46\%$
        & $94.94 \pm 0.25\%$
        & $96.18 \pm 0.18\%$
        \T\B\\
    \hline
        \multicolumn{1}{l|}{{\sl KD-FixMatch-SCE-1.0-0.01}}
        %& \% & \% & \% & \%
        & $87.33 \pm 0.52\%$
        & $93.03 \pm 0.54\%$
        & ${\bf 95.04 \pm 0.18\%}$
        & $96.33 \pm 0.23\%$
        \T\B\\
    \hline
        \multicolumn{1}{l|}{{\sl KD-FixMatch-SCE-1.0-0.1}}
        %& \% & \% & \% & \%
        & ${\bf 87.55 \pm 0.41\%}$
        & ${\bf 93.19 \pm 0.32\%}$
        & $95.00 \pm 0.41\%$
        & ${\bf 96.34 \pm 0.13\%}$
        \T\B\\
    \hline
    \end{tabular}
    }
    \subcaption{{\sf CIFAR10}}
    \end{center}

% SVHN
    \begin{center}
    \resizebox{1.0\columnwidth}{!}{%
    \begin{tabular}{l | c c c c}
    \hline
      & \multicolumn{1}{c}{$400$ labels}
      & \multicolumn{1}{c}{$1,000$ labels}
      & \multicolumn{1}{c}{$2,000$ labels}
      & \multicolumn{1}{c}{$4,000$ labels}
        \T\B\\
    \hline
        \multicolumn{1}{l|}{{\sl Baseline}}
        & $19.66 \pm 23.56\%$
        & $30.23 \pm 26.62\%$
        & $82.97 \pm 0.98\%$
        & $87.06 \pm 0.22\%$
        \T\B\\
    \hline
        \multicolumn{1}{l|}{{\sl FixMatch}}
        & $70.20 \pm 2.11\%$
        & $83.40 \pm 0.85\%$
        & $87.84 \pm 1.90\%$
        & $89.91 \pm 0.88\%$
        \T\B\\
    \hline
        \multicolumn{1}{l|}{{\sl KD-FixMatch-CE}}
        & $76.01 \pm 1.37\%$
        & $84.50 \pm 1.45\%$
        & ${\bf 89.07 \pm 0.90\%}$
        & $91.12 \pm 0.38\%$
        \T\B\\
    \hline
        \multicolumn{1}{l|}{{\sl KD-FixMatch-SCE-1.0-0.01}}
        & $76.46 \pm 1.65\%$
        & ${\bf 84.52 \pm 1.36\%}$
        & $89.06 \pm 1.09\%$
        & ${\bf 91.15 \pm 0.75\%}$
        \T\B\\
    \hline
        \multicolumn{1}{l|}{{\sl KD-FixMatch-SCE-1.0-0.1}}
        & ${\bf 76.50 \pm 1.80\%}$
        & $84.40 \pm 1.62\%$
        & $88.50 \pm 1.08\%$
        & $91.11 \pm 0.45\%$
        \T\B\\
    \hline
    \end{tabular}
    }
    \subcaption{{\sf SVHN}}
    \end{center}

% CIFAR100
    \begin{center}
    \resizebox{1.0\columnwidth}{!}{%
    \begin{tabular}{l | c c c c}
    \hline
      & \multicolumn{1}{c}{$1,000$ labels}
      & \multicolumn{1}{c}{$4,000$ labels}
      & \multicolumn{1}{c}{$10,000$ labels}
      & \multicolumn{1}{c}{$20,000$ labels}
        \T\B\\
    \hline
        \multicolumn{1}{l|}{{\sl Baseline}}
        & $35.02 \pm 30.12\%$
        & $71.92 \pm 0.36\%$
        & $78.20 \pm 0.34\%$
        & $81.97 \pm 0.36\%$
        \T\B\\
    \hline
        \multicolumn{1}{l|}{{\sl FixMatch}}
        & $57.72 \pm 1.02\%$
        & $73.73 \pm 0.53\%$
        & $79.51 \pm 0.87\%$
        & $82.74 \pm 0.30\%$
        \T\B\\
    \hline
        \multicolumn{1}{l|}{{\sl KD-FixMatch-CE}}
        %& \% & \% & \% & \%
        & ${\bf 61.26 \pm 0.96\%}$
        & $75.94 \pm 0.65\%$
        & $81.05 \pm 0.31\%$
        & $83.53 \pm 0.52\%$
        \T\B\\
    \hline
        \multicolumn{1}{l|}{{\sl KD-FixMatch-SCE-1.0-0.01}}
        %& \% & \% & \% & \%
        & $60.68 \pm 0.96\%$
        & ${\bf 76.01 \pm 0.61\%}$
        & $80.75 \pm 0.32\%$
        & $83.65 \pm 0.36\%$
        \T\B\\
    \hline
        \multicolumn{1}{l|}{{\sl KD-FixMatch-SCE-1.0-0.1}}
        %& \% & \% & \% & \%
        & $60.62 \pm 0.96\%$
        & $75.57 \pm 0.13\%$
        & ${\bf 81.10 \pm 0.15\%}$
        & ${\bf 83.85 \pm 0.28\%}$
        \T\B\\
    \hline
    \end{tabular}
    }
    \subcaption{{\sf CIFAR100}}
    \end{center}

% FOOD101
    \begin{center}
    \resizebox{1.0\columnwidth}{!}{%
    \begin{tabular}{l | c c c c}
    \hline
      & \multicolumn{1}{c}{$1,000$ labels}
      & \multicolumn{1}{c}{$4,000$ labels}
      & \multicolumn{1}{c}{$10,000$ labels}
      & \multicolumn{1}{c}{$20,000$ labels}
        \T\B\\
    \hline
        \multicolumn{1}{l|}{{\sl Baseline}}
        & $38.81 \pm 0.38\%$
        & $59.87 \pm 0.42\%$
        & $69.97 \pm 0.19\%$
        & $75.62 \pm 0.17\%$
        \T\B\\
    \hline
        \multicolumn{1}{l|}{{\sl FixMatch}}
        & $39.94 \pm 0.42\%$
        & $63.26 \pm 0.46\%$
        & $72.04 \pm 0.33\%$
        & $77.38 \pm 0.23\%$
        \T\B\\
    \hline
        \multicolumn{1}{l|}{{\sl KD-FixMatch-CE}}
        & $42.64 \pm 0.96\%$
        & $65.60 \pm 0.32\%$
        & $74.72 \pm 0.39\%$
        & $79.33 \pm 0.33\%$
        \T\B\\
    \hline
        \multicolumn{1}{l|}{{\sl KD-FixMatch-SCE-1.0-0.01}}
        & ${\bf 43.18 \pm 1.23\%}$
        & $65.94 \pm 1.08\%$
        & $74.54 \pm 0.30\%$
        & $79.29 \pm 0.24\%$
        \T\B\\
    \hline
        \multicolumn{1}{l|}{{\sl KD-FixMatch-SCE-1.0-0.1}}
        & $41.73 \pm 1.51\%$
        & ${\bf 66.12 \pm 1.10\%}$
        & ${\bf 74.86 \pm 0.33\%}$
        & ${\bf 79.46 \pm 0.09\%}$
        \T\B\\
    \hline
    \end{tabular}
    }
    \subcaption{{\sf FOOD101}}
    \end{center}
    \vspace{-5.0mm}
    \caption{Test accuracy for four public data sets.
        We run all the experiments five times using the same five random seeds and report the mean and standard deviation of the test accuracy.
        %In order to have the fair comparison, for each column, we use the same labeled train data for the same random seed in all the compared methods.
        The best results in each column are represented in bold letters.
    }
\label{table:exp0}
\end{table}

\section{Conclusion}
\label{sec:conclusion}

In this paper, we have proposed an SSL algorithm, {\sl KD-FixMatch}, which is an improved version of {\sl FixMatch} that utilizes knowledge distillation.
Based on our experiments, {\sl KD-FixMatch} outperforms {\sl FixMatch} and {\sl Baseline} on four public data sets.
Interestingly, despite the absence of the parameter selection for SCE (robust loss function), the performance of {\sl KD-FixMatch-SCE} is slightly better than {\sl KD-FixMatch-CE} in most of the cases.

\bibliographystyle{IEEEbib}
\bibliography{sdp}

\end{document}